\documentclass[10pt, logo, twocolumn, copyright]{nvidiatechreport}

\usepackage[numbers,sort&compress]{natbib}

\titlespacing*{\paragraph}{0pt}{0.5ex plus 0.0ex minus 0.5ex}{1em}

\title{From Modalities to Propositions: A Language-Centric Framework for Multimodal Intelligence}
\author[1]{Nadine Chang*}
\author[1]{Maying Shen*}
\author[1]{Shizhe Diao}
\author[1]{Jialiang Wang}
\author[1]{Jingde Chen}
\author[1]{Thomas Breuel}
\author[1]{Pavlo Molchanov}
\author[1,2]{Rafid Mahmood}
\author[1]{Jose Alvarez}
\affil[1]{NVIDIA}
\affil[2]{University of Ottawa}

\usepackage[utf8]{inputenc} 
\usepackage[T1]{fontenc}    
\usepackage{hyperref}       
\usepackage{url}            
\usepackage{booktabs}       
\usepackage{amsfonts}       
\usepackage{nicefrac}       
\usepackage{microtype}      
\usepackage[dvipsnames]{xcolor}         
\usepackage{arydshln}

\usepackage{hyperref}
\hypersetup{
    colorlinks=true,    
    citecolor=citeblue, 
    linkcolor=red,      
    urlcolor=blue       
}

\definecolor{citeblue}{RGB}{0, 114, 189} 

\usepackage{graphicx}
\usepackage{booktabs}
\usepackage{xspace}
\usepackage{csquotes}
\usepackage{color, colortbl}
\usepackage{enumitem}
\usepackage{subcaption}
\usepackage[ruled,linesnumbered]{algorithm2e}

\usepackage{amsmath}
\usepackage{makecell}
\usepackage{multirow}
\usepackage{float}
\usepackage{amssymb}
\usepackage{utfsym}
\usepackage{diagbox}
\usepackage{dsfont}
\usepackage{cleveref}
\crefname{section}{Sec.}{Secs.}
\Crefname{section}{Section}{Sections}
\Crefname{table}{Table}{Tables}
\crefname{table}{Tab.}{Tabs.}
\crefname{algocf}{Alg.}{Algs.}
\Crefname{algocf}{Algorithm}{Algorithms}

\PassOptionsToPackage{numbers, compress}{natbib}

\newcommand{\para}[1]{\medskip\noindent\textbf{#1.}}

\definecolor{queryAction}{HTML}{DCEBFA}
\definecolor{queryStatus}{HTML}{F8D7DA}
\definecolor{queryWeather}{HTML}{DDF3E4}
\definecolor{queryAnimal}{HTML}{EEE2FA}

\DeclareRobustCommand{\queryblock}[2]{%
    \begingroup
    \setlength{\fboxsep}{3pt}%
    \colorbox{#1}{\strut\textbf{#2}}%
    \endgroup
}

\DeclareRobustCommand{\queryplus}{%
    \hspace{0.4em}\textbf{+}\hspace{0.4em}%
}

\begin{abstract}
We propose a language representation for multimodal data in which any observation, whether image, video, or text, is expressed as a bag of atomic propositions, simple statements about the entities, actions, and relations in a scene. A global semantic codebook unifies these into a shared vocabulary of canonical atomic propositions, placing every modality and observation into one interpretable space that spans fine grained facts to high level concepts and composes into richer ones. This brings interpretability with reasoning, cross-modal understanding and retrieval, and compositionality that enables complex multimodal understanding, rich data curation and complex structured retrieval. We demonstrate the framework on autonomous driving and open-world data.

\end{abstract}

\begin{document}
\maketitle

\begin{figure*}[!t]
    \centering
    \includegraphics[width=1.0\textwidth]{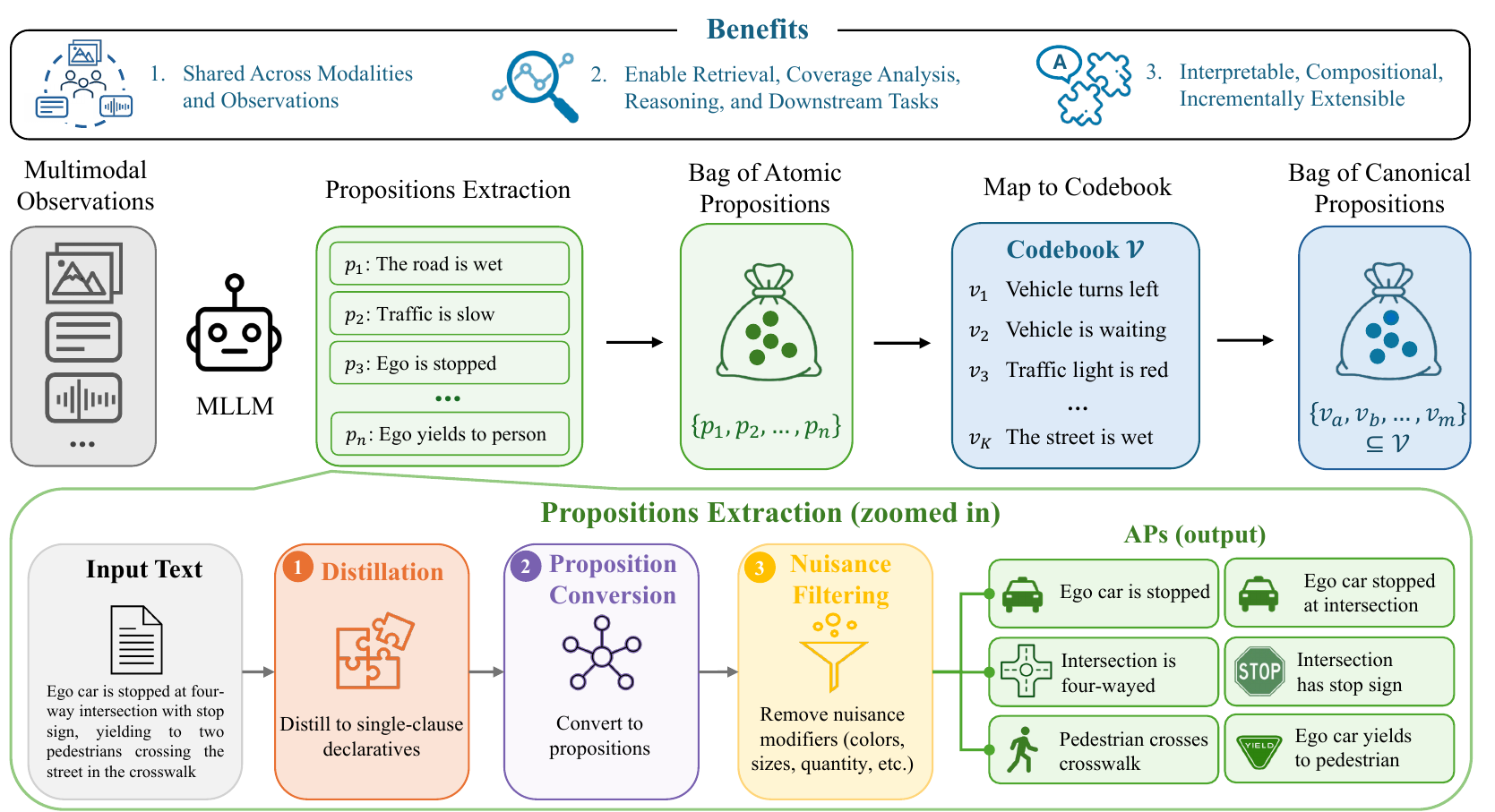}
    \caption{\textbf{Overview of our proposition based semantic representation pipeline.} Top: Benefits of atomic proposition (AP) representations. Middle: Framework overview. An observation, is processed by a MLLM to extract a set of natural language APs describing the scene, objects, actions, and events. These extracted APs form a bag of APs, capturing the semantic content of the observation. Each AP is then mapped to its closest canonical entry in a global semantic codebook, a vocabulary of standardized APs shared across observations and modalities. The resulting bag of canonical APs provides a compact, interpretable, and compositional representation that supports coverage analysis, reasoning, retrieval, and downstream ML tasks while staying extendable as new concepts are introduced. Bottom: AP extraction pipeline from observations.
    }
    \label{fig:overview}
\end{figure*} 

\section{Introduction}

Learning to represent multimodal data is the fundamental component for modern AI. Vision language models and multimodal language models (MLLMs) learn shared embedding spaces that capture rich semantic information across images, text, audio, and video, enabling strong performance on recognition, retrieval, generation, and reasoning tasks~\cite{radford2021learning,liu2023visual,bai2023qwenvl,liang2026comprehensive}.
Despite their success, these representations are largely opaque~\cite{bommasani2021opportunities,kazmierczak2025explainability}. Semantic concepts are distributed across high dimensional embeddings, making it difficult to identify relevant information, interpret model behavior, or perform structured reasoning~\cite{elhage2022toy,bereska2024mechanistic}. This limitation is important because real world understanding is inherently compositional, involving entities, attributes, actions, and relationships that combine into higher level semantics~\cite{thrush2022winoground,changposition}, and a single representation muddles this richness.

Inspired by the classic bag-of-words (BoW) representation in natural language processing~\cite{manning1999foundations,jurafsky2009speech}, we propose propositions as an atomic unit to represent observed data samples from any modality, such as an image, video, text, or audio. We define of Bag of Atomic Propositions (BoAP) as a collection of elementary semantic statements such as \textit{"person holding cup"}, \textit{"dog chasing ball"}, or \textit{"car parked beside building"}. Similar to how BoW represents a document by the words it contains, BoAP represents an observation by the atomic propositions it evokes, unifying all modalities within a single, shared semantic vocabulary.

Decomposing an observation into these reusable semantic primitives exposes its compositional structure. Specifically, complex situations can be expressed as combinations of simple concepts.
Thus, higher level concepts can be composed directly from our interpretable atomic propositions, enabling explicit compositional reasoning rather than relying on reasoning through the implicit relationships buried in a dense embedding space~\cite{cai2025naver,morishita2024enhancing}. Instead of replacing dense embeddings, BoAP serves as a complement. Embeddings capture fine-grained perceptual information but remain opaque, while APs expose explicit semantic structure that can be inspected and composed. Combining them keeps the perceptual robustness of embeddings while adding the interpretability and compositionality of a symbolic representation.

In this paper, we introduce a framework that leverages MLLMs to extract atomic propositions (APs) from multimodal observations and process APs to address semantically similar and redundant ones. Because the same concept can be phrased in many ways, we normalize equivalent APs to a single canonical form. This yields a global semantic codebook, a shared vocabulary of canonical atomic propositions (APs). ~\cref{fig:overview} shows an overview of our framework.

We demonstrate the benefits of our AP representations for autonomous driving and open-world videos. Both settings have diverse agents, environments, infrastructure, and conditions that combine into complex behaviors and interactions. We focus on three benefits. The first is a \emph{description unification} of any modality and observation in a single shared propositional vocabulary. Consequently, semantically equivalent situations map to the same APs regardless of their data source. The second is \emph{compositionality}, the ability to express and reason about complex situations as combinations of APs. The third is the ability to identify \emph{knowledge coverage}. By representing both a training set and an evaluation set as bags of APs, we identify the conceptual knowledge gap a model must handle at deployment but has rarely or never seen during training. Because APs are compositional, we surface these gaps across a spectrum, ranging from highly fine grained gaps corresponding to individual APs to highly complex gaps corresponding to combinations of APs. Quantifying coverage in this way enables targeted data collection, data mixing, and synthetic data generation for uncollectable data, and ultimately more robust models.

\section{From Observations to Propositions: A Compositional Framework}
Our framework consists of two complementary components. The first extracts semantic atomic propositions from individual multimodal observations. The second constructs a global semantic codebook that unifies semantically equivalent APs across the entire dataset. Together, these components transform heterogeneous multimodal inputs into a shared, interpretable semantic representation.

~\cref{fig:overview} illustrates the full pipeline. We start from an observation in any modality, such as a video, image, text log, or sensor stream. A multimodal large language model (MLLM) first describes the observation in natural language, and we extract from this description a set of APs, producing a bag of APs (BoAP) for the observation. We then map each AP to its canonical form in a global semantic codebook, a vocabulary of canonical APs shared across the whole corpus. The result is a bag of canonical APs that places every observation in one common, interpretable semantic space. The remainder of this section details each component. Section~\ref{sec:extraction} describes how APs are extracted from a single observation, and Section~\ref{sec:codebook} describes how the codebook is built across the corpus.

\subsection{Proposition Extraction from Multimodal Inputs}
\label{sec:extraction}

Consider a multimodal observation $x$ spanning several modalities such as images, video, text, audio, or sensor streams. Our goal is to extract a set of APs that captures its semantic content, where each AP is an elementary semantic statement describing an entity, attribute, action, event, or relation in the observation. 
This representation preserves semantically salient information while discarding nuisance details that do not contribute to scene understanding.

Our extraction pipeline takes as input a natural language description of the observation. As shown in Figure \ref{fig:overview}, it consists of three stages: (i) distillation, which reduces the description to a set of statements; (ii) proposition conversion, which converts these statements into APs; and (iii) nuisance filtering, which removes task-irrelevant modifiers. When operating over a dataset, the resulting APs are aligned across all observations so that semantically equivalent concepts share a common representation.

\subsubsection{Distillation}

Distillation transforms the input description into a set of simple, single-clause declarative sentences, each expressing one single fact. This step removes narrative and stylistic complexity and separates nested clauses while preserving the salient content. For example, the caption \textit{``A pedestrian is crossing a wet street while a vehicle waits at a red light''} is distilled into \textit{``a pedestrian is crossing the street''}, \textit{``the street is wet''}, \textit{``a vehicle is waiting''}, and \textit{``the traffic light is red''}. We perform distillation with a single Gemini-2.5-Pro call per description, using the prompt in the appendix (~\cref{fig:distillation_prompt}). A single model is efficient and works well in practice, and the stage can be made more robust by ensembling or contrasting multiple LLMs when compute permits.

\subsubsection{Proposition Conversion}
\label{sec:conversion}

Proposition conversion turns each distilled sentence into one or more APs. An AP is the smallest self contained statement that expresses a single fact, capturing one entity, attribute, action, event, or relation that cannot be split further without losing meaning. Each AP is produced and processed by an LLM, and we treat it as a discrete semantic token that can be reused across observations, modalities, and domains. Formally, we write an AP as
\begin{equation}
p=(r,a_1,\ldots,a_k),
\end{equation}
\noindent where $r$ is a relation or predicate and the $a_i$ are its noun-anchored arguments, the entities or attributes it relates. The observation as a whole is then represented by its set of APs,
\begin{equation}
P(x)=\{p_1,p_2,\ldots,p_n\}.
\end{equation}

We generate APs with an LLM using a structured decomposition prompt. The prompt rewrites each distilled sentence into one or more APs by expanding conjunctions into separate APs and resolving pronouns and references to the entities they name.

In practice, we implement this with Gemini-2.5-Pro via two pass prompting. First, we identify the salient entities and then the subject, predicate, object relations among them, with subjects and objects constrained to the extracted entity list. The output for each observation is a JSON set of triplets, each treated as an AP.

Continuing the running example, the distilled sentences \textit{``a pedestrian is crossing the street''}, \textit{``the street is wet''}, \textit{``a vehicle is waiting''}, and \textit{``the traffic light is red''} are converted into the AP set
\begin{equation}
\begin{split}
P(x)=\{&\text{pedestrian crossing street},\ \text{street is wet}, \\
&\text{vehicle waiting},\ \text{traffic light red}\}.
\end{split}
\end{equation}
Each AP can further be assigned to a semantic category, such as weather, operational design domain, agent behavior, sensor state, agent interaction, or road infrastructure, so that APs can be systematically organized and reasoned over. For autonomous driving, we group them under a typed ontology of Actor, Action, Environment, Infrastructure, Interaction, and Intent. 
Finally, note that unlike scene captions or free form descriptions, APs are independent and compositional. Consequently, the same AP can recur across many observations and modalities and be aggregated at the corpus level.

\begin{figure*}[!t]
    \centering
    \includegraphics[width=1.0\textwidth]{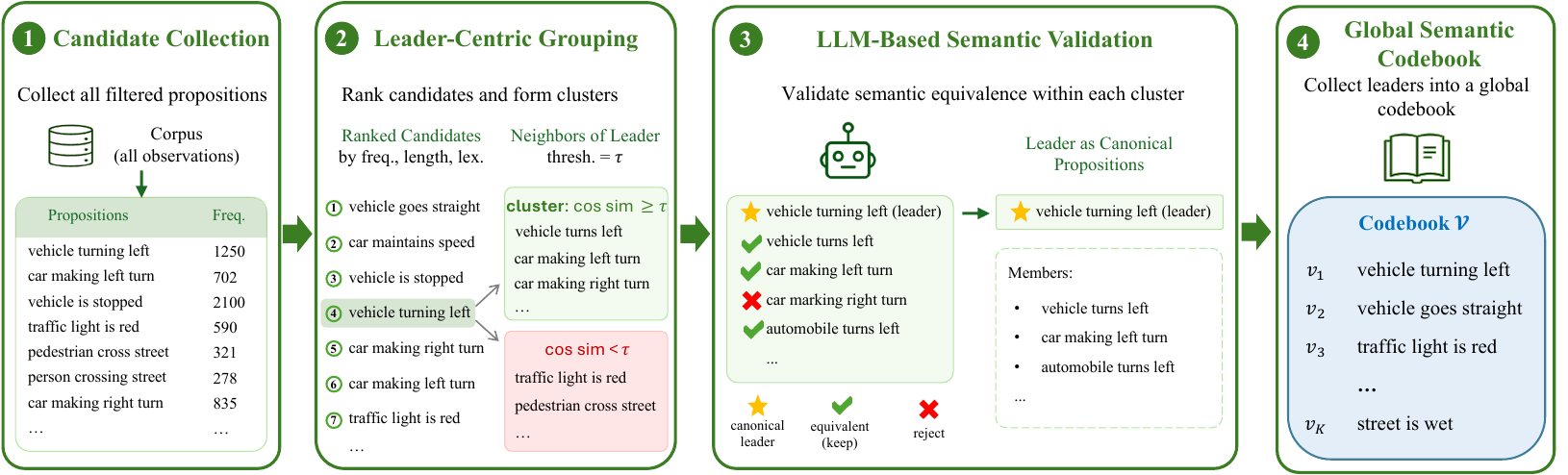}
    \caption{\textbf{Global semantic codebook construction for canonical AP generation.} Processed APs extracted from all observations in the corpus are first aggregated and ranked according to frequency, phrase length, and lexical ordering. A leader-centric grouping strategy then iteratively selects high ranking APs as group leaders and identifies semantically similar neighbors using embedding based similarity. Possible similar APs are subsequently validated by an LLM, which verifies whether these APs are semantically equivalent to the leader and rejects incorrectly assigned APs. Validated groups are consolidated into canonical clusters, where the leader serves as the canonical AP representing the shared semantic concept. The resulting set of cluster leaders forms a global semantic codebook, providing a corpus wide vocabulary of canonical APs that can be used to map observation level APs into a unified and interpretable semantic representation.}
    \label{fig:canonical}    
\end{figure*} 

\subsubsection{Nuisance Filtering}

Proposition conversion favors completeness, so the extracted APs often carry modifiers that are irrelevant to the task, such as colors, sizes, and quantities. The difficulty is that what counts as a nuisance is itself task specific and not obvious to define. The same attribute can be noise for one task yet essential for another. Color, for instance, is usually irrelevant, but it matters when the task concerns a red traffic light or an emergency vehicle. Deciding what to remove therefore requires task knowledge, and we treat our default set of modifiers, namely color, size, and quantity, as a pragmatic and configurable choice rather than a universal rule.

Nuisance filtering has two goals. First, it removes nuisance modifiers from AP. Second, it normalizes minute variation such as articles, plurality, and proper nouns (\eg specific car brands). Together these remove a large amount of lexical duplication. We collect all entity arguments across the extracted APs and use Gemini-2.5-Pro to strip non essential modifiers (prompt in appendix (~\cref{fig:filtering_prompt}). We apply this denoising only to entity arguments and leave predicate components in their original form.

\subsection{Global Semantic Codebook Construction}
\label{sec:codebook}

The same concept is often phrased differently. For example, \textit{``pedestrian crossing street''}, \textit{``person crossing the road''}, and \textit{``pedestrian walking across the street''} all denote the same type of event. This redundancy fragments the proposition space and weakens downstream tasks such as retrieval, coverage analysis, and representation learning. To remove this redundancy, we build a global semantic codebook that maps every AP to a canonical form shared across the corpus.

Algorithm \ref{alg:prop-dedup} details the steps of our procedure summarized in Figure \ref{fig:canonical}. First, candidate collection ranks all propositions by their occurrence frequency, specificity and lexicographic order. Second, leader-centric grouping then uses embedding similarity to gather, around each frequent AP, a candidate set of possible variants. Third, LLM-based verification checks which candidates truly share its meaning. Finally, the surviving representatives form the canonical codebook. The first two steps favor recall, gathering likely variants cheaply, while verification enforces precision, so that only genuinely equivalent APs are merged. The remainder of this section details each step in turn.

\subsubsection{Candidate Collection}
Let
\begin{equation}
\mathcal{P} = \bigcup_{x \in \mathcal{C}} {{P}}(x),    
\end{equation}
\noindent denote the multiset of all filtered APs extracted from the corpus $\mathcal{C}$. 
Candidate collection turns this pool into a single ranked list, so that the most frequent concepts are considered first when the codebook is formed.
For each unique proposition $p_i\in \mathcal{P}$, we compute its corpus frequency,
\begin{equation}
f(p_i)=\mid \{x \in \mathcal{C}:p_i \in P(x)\}\mid,    
\end{equation}
\noindent which we use as a measure of semantic prevalence across observations. We then order candidate APs according to three hierarchical criteria:
\begin{itemize}
    \item Corpus frequency (descending), prioritizing concepts that occur most frequently across observations;
    \item Proposition conciseness (ascending AP length), favoring shorter, more canonical phrasings when frequencies are identical;
    \item Lexicographic order, providing deterministic tie-breaking and reproducibility.
\end{itemize}

This ranking strategy establishes a stable ordering over the proposition space and ensures that frequently occurring semantic concepts are preferentially selected as cluster leaders during subsequent codebook induction. The resulting ranked repository serves as the input to the leader-centric grouping stage, where semantically redundant propositions are consolidated into canonical APs.

\subsubsection{Leader-Centric Grouping}

Leader-centric grouping gathers candidate variants that denote the same concept around the most frequent APs. Working down the ranked list, we take the highest ranked unassigned AP as a leader $l$. We embed every proposition with a language embedding model (Qwen3-Embedding-8B) and attach to $l$ each remaining unassigned AP whose cosine similarity to $l$ is at least a threshold $\tau$ in line \ref{alg:line9}.
For example, the leader \textit{``pedestrian crossing street''} attracts variants such as \textit{``person crossing the road''} and \textit{``pedestrian walking across the street''}. The leader and its candidate variants are then removed from the unassigned pool, and we repeat with the next leader until the pool is empty (the grouping loop of ~\cref{alg:prop-dedup}). Because leaders are chosen by frequency, common concepts become the representatives and rarer phrasings collapse onto them.

\subsubsection{LLM-Based Semantic Equivalence Verification} 

Embedding similarity efficiently proposes candidates, but high similarity does not guarantee that two APs share a meaning. For example, \textit{``vehicle stopping''}, \textit{``vehicle parking''}, and \textit{``vehicle yielding''} sit close in embedding space yet describe different behaviors. We therefore verify each candidate set with an LLM.

We present the leader $l$ and its candidates $\mathcal{N}(l)$ to the LLM together with domain instructions that define equivalence for autonomous driving (AD) (appendix ~\cref{fig:verification_prompt}), and ask which candidates are truly equivalent to $l$. Equivalent APs are assigned to $l$ group, and the rest return to the unassigned pool, where they may later become leaders or join other groups. Embedding search thus acts as a high recall candidate generator and the LLM as a high precision verifier.

We note that our method can also be applied at a finer grain, specifically to the entity and predicate components of the APs separately. This gives users a second option, canonicalizing at the component level rather than over whole APs, while we use whole proposition canonicalization as the default.

\IncMargin{1.5em}
\begin{algorithm}[t]

\caption{Global Semantic Codebook Construction}
\label{alg:prop-dedup}
\KwIn{
Processed propositions $\mathcal{P}$;
similarity function $s(\cdot,\cdot)$; similarity threshold $\tau$
}
\KwOut{Canonical proposition vocabulary $\mathcal{V}$}

$\mathcal{V} \leftarrow \emptyset$\;

\BlankLine
\tcp{Candidate Collection}
\For{$p_i\in\mathcal{P}$}{
    calculate frequency $f(p_i)$
}
Sort $\mathcal{P}$ by descending
$f(\cdot)$, phrase length, and lexicographic order\;

\BlankLine
\tcp{Leader-Centric Grouping}
$\mathcal{U} \leftarrow \mathcal{P}$ \tcp*[r]{unassigned propositions}
\While{$\mathcal{U} \neq \emptyset$}{
    $l \leftarrow$ highest ranked $p_i$ from $\mathcal{U}$\;

    $\mathcal{N}(l) \leftarrow
    \{p_j \in \mathcal{U} : p_j \neq l \land s(l,p_j) \geq \tau\}$\; \label{alg:line9}

    \BlankLine
    \tcp{Semantic Equivalence Verification}
    $\mathcal{S} \leftarrow \mathrm{LLMVerify}(l, \mathcal{N}(l))$
    \tcp*[r]{verified equivalents of $l$}

    $\mathcal{V} \leftarrow \mathcal{V} \cup \{l\}$\;
    $\mathcal{U} \leftarrow \mathcal{U} \setminus (\{l\} \cup \mathcal{S})$\;
}

\Return{$\mathcal{V}$}\;
\end{algorithm}

\subsubsection{Global Semantic Codebook}
Following semantic validation, each validated cluster represents a distinct semantic concept discovered from the corpus. The cluster leader serves as the canonical AP, $\nu_i$, for that concept, while the remaining cluster members correspond to alternative lexical phrasings that have been verified as semantically equivalent. We define the collection of all canonical APs, $\mathcal{V}$, as a global semantic codebook.


This codebook is analogous to a visual vocabulary in bag-of-words systems or a codebook in vector quantization. However, unlike conventional codebooks constructed from low level feature descriptors, the proposed codebook is composed of human interpretable semantic APs that explicitly represent entities, actions, attributes, and relations.

Each AP extracted from an observation can be subsequently mapped to its corresponding canonical AP through a codebook assignment function:
\begin{equation}
    g:p_i \to \nu_j,
\end{equation}
\noindent where $p_i$ is an extracted AP and $\nu_j \in \mathcal{V}$ is the canonical AP associated with the cluster containing $p_i$.

Importantly, this mapping transforms observation specific proposition sets $P(x)=\{p_1,p_2, \dots , p_n\} $ into canonical semantic representations,
\begin{equation}
    \mathcal{B}(x) =\{ g(p_1),g(p_2),\dots ,g(p_n) \} \subseteq \mathcal{V}.
\end{equation}

 \begin{figure*}[!ht]
    \centering

    \begin{subfigure}{1.0\textwidth}
        \centering
        \includegraphics[width=1.0\textwidth]{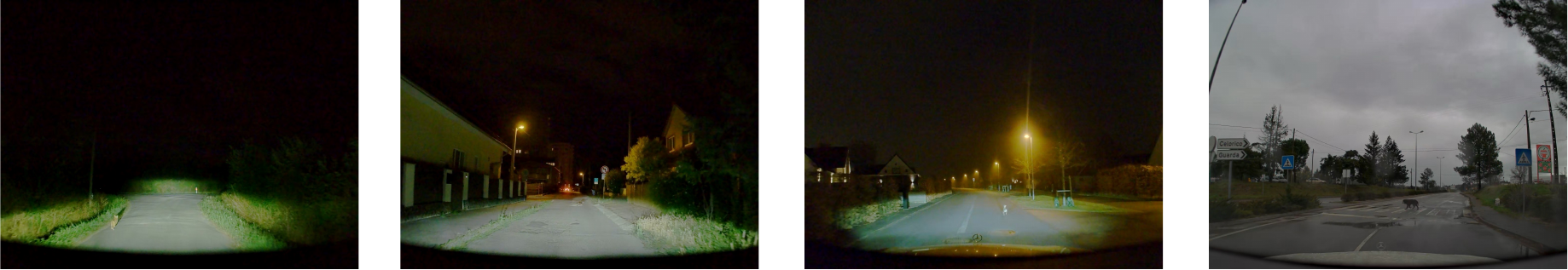}
        \caption*{
            \queryblock{queryAction}{animal crossing the road}
        }
    \end{subfigure}

    \caption{\textbf{BoAP scenario retrieval with a single query.} Retrieved examples for \textit{animal cross the road} with scenarios. We retrieve accurate scenarios across different time, environment, and object type. }
    \label{fig:animal_query}
\end{figure*}
\begin{figure*}[!t]
    \centering
    
    \begin{subfigure}{1.0\textwidth}
        \centering
        \includegraphics[width=1.0\textwidth]{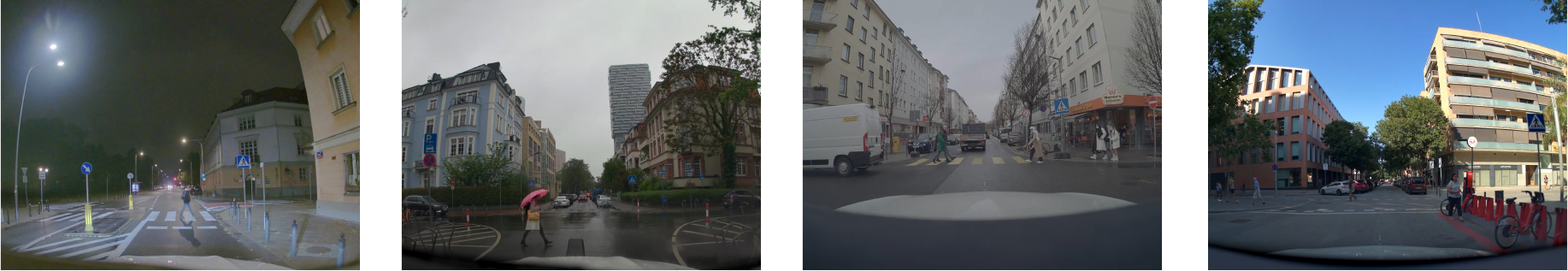}
        \caption*{
            \queryblock{queryAction}{pedestrian crossing street}
        }
        \label{fig:composite_query_single}
    \end{subfigure}

    \vspace{1em}

    \begin{subfigure}{1.0\textwidth}
        \centering
        \includegraphics[width=1.0\textwidth]{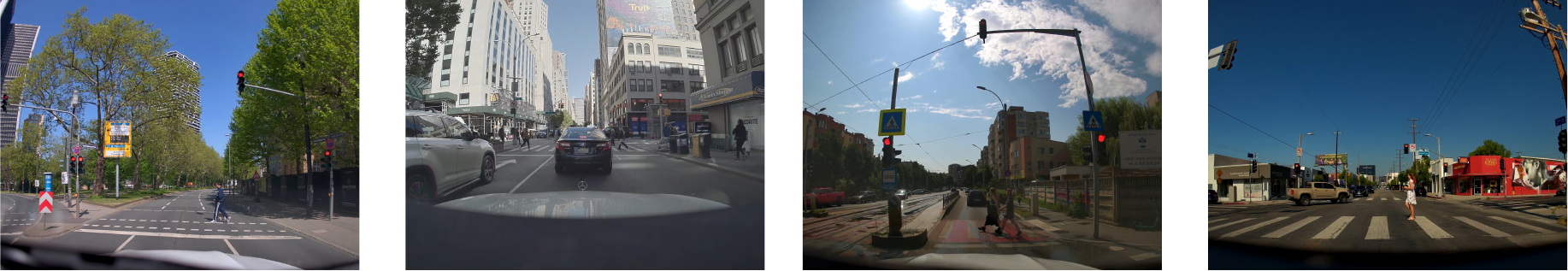}
        \caption*{
            \queryblock{queryAction}{pedestrian crossing street}
            \queryplus
            \queryblock{queryStatus}{red traffic light}
        }
        \label{fig:composite_query_double}
    \end{subfigure}

    \vspace{1em}

    \begin{subfigure}{1.0\textwidth}
        \centering
        \includegraphics[width=1.0\textwidth]{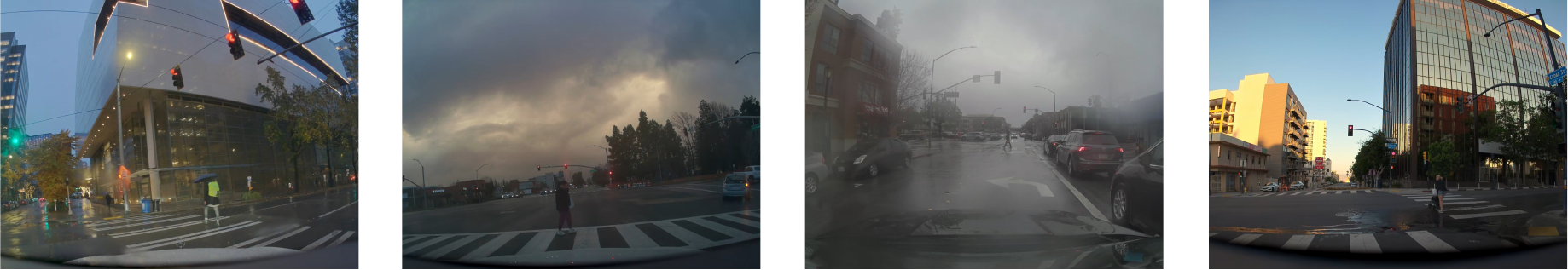}
        \caption*{
            \queryblock{queryAction}{pedestrian crossing street}
            \queryplus
            \queryblock{queryStatus}{red traffic light}
            \queryplus
            \queryblock{queryWeather}{road is wet}
        }
        \label{fig:composite_query_complex}
    \end{subfigure}
    
    \caption{\textbf{Physical-AI: BoAP composite query retrieval scenarios.} We progressively increase our query complexity and visually show at each stage, accurate composite retrievals across a wide spectrum of time, environment, infrascture, and agents. }
    \label{fig:composite_query}
\end{figure*} 
\begin{figure*}[!t]
    \centering

    \captionsetup[subfigure]{
        justification=centering,
        singlelinecheck=false
    }

    \begin{subfigure}{1.0\textwidth}
        \centering
        \includegraphics[width=1.0\textwidth]
            {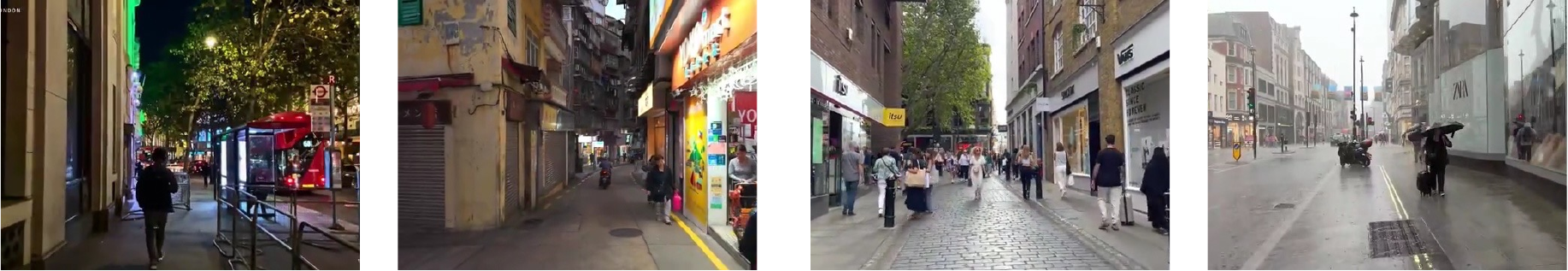}
        \caption*{
            \queryblock{queryAction}{camera moves through urban area}
        }
        \label{fig:sekai_query_single}
    \end{subfigure}

    \vspace{1em}

    \begin{subfigure}{1.0\textwidth}
        \centering
        \includegraphics[width=1.0\textwidth]
            {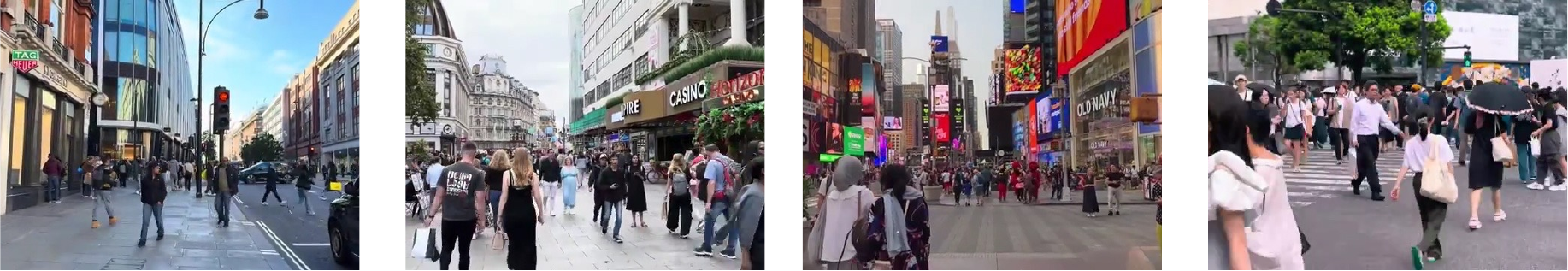}
        \caption*{
            \queryblock{queryAction}{camera moves through urban area}
            \queryplus
            \queryblock{queryStatus}{street is bustling}
        }
        \label{fig:sekai_query_double}
    \end{subfigure}

    \vspace{1em}

    \begin{subfigure}{1.0\textwidth}
        \centering
        \includegraphics[width=1.0\textwidth]
            {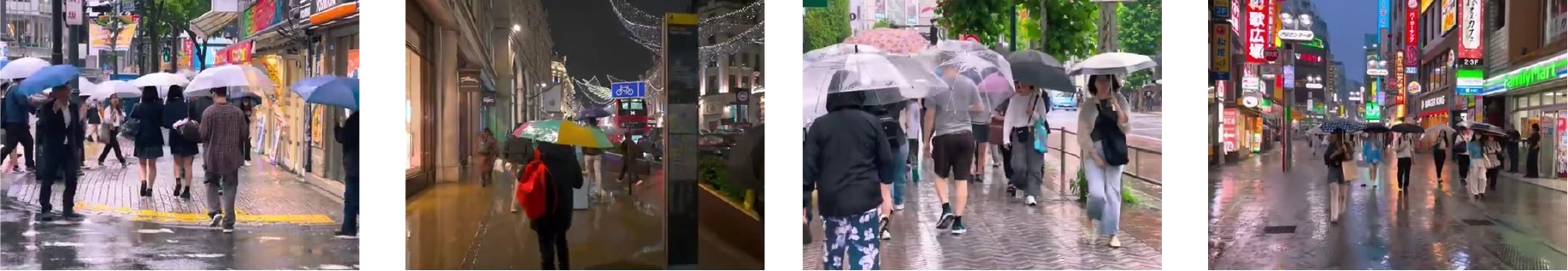}
        \caption*{
            \queryblock{queryAction}{camera moves through urban area}
            \queryplus
            \queryblock{queryStatus}{street is bustling}
            \queryplus
            \queryblock{queryWeather}{weather is rainy}
        }
        \label{fig:sekai_query_multi}
    \end{subfigure}

    \caption{\textbf{Sekai: BoAP composite query retrieval scenarios.} We progressively increase our query complexity and visually show at each stage, accurate composite retrievals across a wide spectrum of geolocations, crowd density, and actions. }
    \label{fig:sekai_queries}
\end{figure*}

For the running example, this replaces each proposition with its canonical form. A variant such as \textit{``a vehicle waiting''} maps to the canonical AP \textit{``vehicle is waiting''}, so the pedestrian scene is represented by the bag of canonical APs $\mathcal{B}(x)=\{$\textit{pedestrian crossing street}, \textit{street is wet}, \textit{vehicle is waiting}, \textit{traffic light is red}$\}$.

Thus, the codebook establishes a shared semantic coordinate system across the entire corpus, enabling direct comparison between observations, efficient retrieval of semantically related scenarios, and knowledge coverage analysis. Because all observations are represented using the same vocabulary, higher level analyses can be performed at the concept level.

\section{Experiments}
We show that the Bag of Atomic Propositions (BoAP) representation enables two capabilities that dense embeddings do not. First, it supports compositional reasoning and semantic retrieval, where complex queries are built and matched as combinations of atomic propositions (APs). Second, the induced set of APs makes the semantic content of a corpus measurable, enabling interpretable coverage and diversity analysis. Below we detail the datasets, models, and metrics.

\para{Datasets}
We apply BoAP on three complementary video corpora spanning AD and open-world exploration.
\textbf{Physical AI.}
We use videos from the NVIDIA PhysicalAI--Autonomous Vehicles dataset~\citep{nvidia2026physicalai}, a geographically diverse collection of real world driving data. The full dataset contains 1,700 hours of planned data-collection drives across 25 countries and more than 2,500 cities, organized into 306,152 20-second clips. It captures diverse traffic patterns, weather conditions, road infrastructure, obstacles, and vulnerable road users. In our experiments, this corpus provides real world driving scenarios for demonstrating fine-grained semantic retrieval and analyzing corpus coverage.
\textbf{NavHard.}
NavHard is the challenging evaluation split of NAVSIM-v2~\citep{dauner2024navsim,cao2025pseudo}, constructed from real-world OpenScene driving logs. Compared with nominal driving splits, it emphasizes operationally difficult traffic interactions and road geometries across Boston, Pittsburgh, Las Vegas, and Singapore. We analyze NavHard together with its companion NavTrain split to characterize long-tailed behavior distributions and identify interpretable train--test coverage gaps.
\textbf{Sekai.}
Sekai~\citep{li2025sekai} is a worldwide video dataset for visual world exploration, containing more than 5,000 hours of walking and drone-view videos from over 100 countries and regions and 750 cities. It spans diverse scenes, weather conditions, crowd densities, activities, and camera motions. We use Sekai to test whether BoAP supports compositional retrieval beyond AD, including queries combining entities, activities, and environmental context.

\textbf{Models.}
We use Gemini-2.5-Pro~\cite{comanici2025gemini} for all the processes during the semantic representation pipeline, specifically AP extraction and LLM semantic validation for codebook construction. For encoding AP text into embeddings, we utilize Qwen3-Embedding-8B~\cite{zhang2025qwen3} for our leader-centric grouping and data retrieval experiments. During t-SNE visualization of semantic coverage, we use the lightweight BGE-M3 embedder~\cite{chen2024m3} to generate semantic embeddings for each data point to ensure computational efficiency.

\begin{table*}[th]
\centering
\small
\caption{\textbf{Similarity score comparison between holistic caption retrieval vs. compositional AP retrieval.} As query complexity ($N$) increases, atomic AP retrieval preserves high similarity across constituent concepts, whereas holistic caption retrieval scores remain stagnant. Bold numbers indicate the best overall score for each complexity level.}
\label{tab:similarity_scores}
\resizebox{\textwidth}{!}{
\begin{tabular}{lllcccc}
\toprule
\textbf{Complexity} & \textbf{Method} & \textbf{Query Text / Components} & \textbf{Video 1} & \textbf{Video 2} & \textbf{Video 3} & \textbf{Video 4} \\ 
\midrule

\multirow{2}{*}{\begin{tabular}[c]{@{}l@{}}\textbf{Low}\\ ($N=1$)\end{tabular}} 
& Caption & pedestrian crossing street & 0.55 & 0.53 & 0.52 & 0.50 \\
& AP & pedestrian crossing street & \textbf{0.94} & \textbf{0.93} & \textbf{0.86} & \textbf{0.88} \\

\midrule

\multirow{4}{*}{\begin{tabular}[c]{@{}l@{}}\textbf{Medium}\\ ($N=2$)\end{tabular}} 
& Caption & pedestrian crossing street, red traffic light & 0.59 & 0.58 & 0.54 & 0.52 \\ \cmidrule{2-7}
& \multirow{2}{*}{AP} 
  & (1) pedestrian crossing street & 0.86 & 0.93 & 0.96 & 0.91 \\
& & (2) red traffic light           & 0.91 & 0.91 & 0.91 & 0.88 \\ \cmidrule{2-7}
& \textbf{AP Mean} & \textbf{Combined Score} & \textbf{0.89} & \textbf{0.92} & \textbf{0.94} & \textbf{0.86} \\

\midrule

\multirow{5}{*}{\begin{tabular}[c]{@{}l@{}}\textbf{High}\\ ($N=3$)\end{tabular}} 
& Caption & pedestrian crossing street, red traffic light, road is wet & 0.61 & 0.59 & 0.59 & 0.56 \\ \cmidrule{2-7}
& \multirow{3}{*}{AP} 
  & (1) pedestrian crossing street & 0.96 & 0.96 & 0.94 & 0.86 \\
& & (2) red traffic light           & 0.88 & 0.88 & 0.86 & 0.88 \\
& & (3) road is wet                 & 0.98 & 0.91 & 0.95 & 0.91 \\ \cmidrule{2-7}
& \textbf{AP Mean} & \textbf{Combined Score} & \textbf{0.94} & \textbf{0.92} & \textbf{0.92} & \textbf{0.88} \\

\bottomrule
\end{tabular}
}
\end{table*}

\subsection{Semantic Reasoning and Querying}

We first show that representing observations as APs supports semantic retrieval and reasoning capabilities beyond dense embeddings. Each observation is a bag of APs drawn from the global semantic codebook, and queries are posed directly in AP space, either as a single concept (e.g., \textit{pedestrian crossing street}) or as a composition of several (e.g., \textit{pedestrian crossing street} AND \textit{traffic light red} AND \textit{wet road}).

\subsubsection{Single-Proposition Retrieval and Interpretability}

At the simplest level, an observation is retrieved when its bag of APs contains the queried concept. Retrieval is therefore matching over interpretable propositions rather than a nearest neighbor search over dense embeddings of long descriptions. Prior work has shown the benefit of decomposing captions and queries into finer units~\cite{yang2026scalable}. APs extend this by making each concept a reusable, matchable unit. Fig.~\ref{fig:animal_query} shows single AP retrieval, where an isolated concept such as \textit{animal crossing the road} maps directly to its matching scenarios. This matching provides an interpretable foundation for compositional queries.

\subsubsection{Compositional Generalization in Semantic Space}

\begin{figure}[!h]
\centering
    \begin{subfigure}[b]{0.48\columnwidth}
        \centering
        \includegraphics[width=\textwidth]{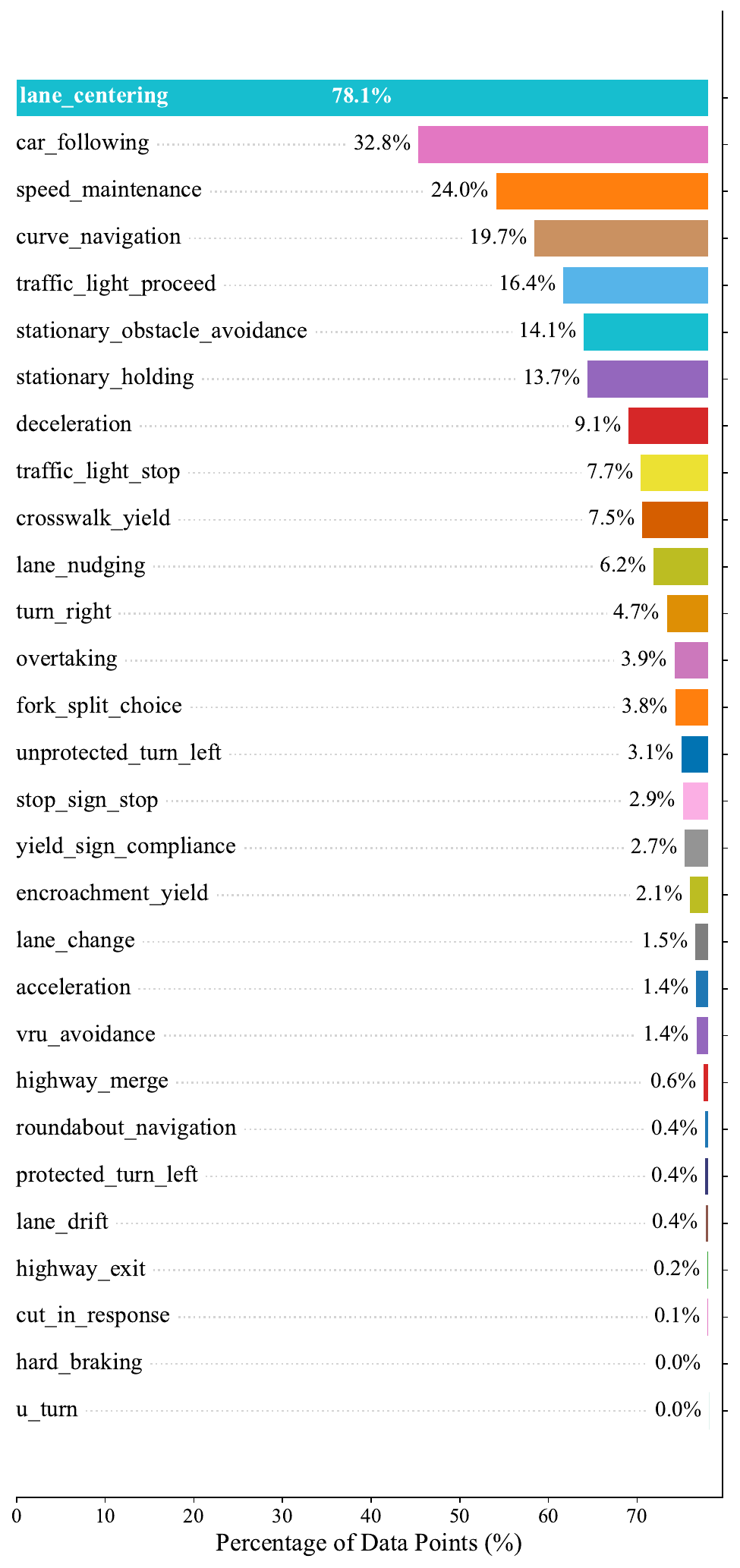}
        \caption{NavHard dataset}
        \label{fig:navhard_distribution}
    \end{subfigure}
    \begin{subfigure}[b]{0.48\columnwidth}
        \centering
        \includegraphics[width=\textwidth]{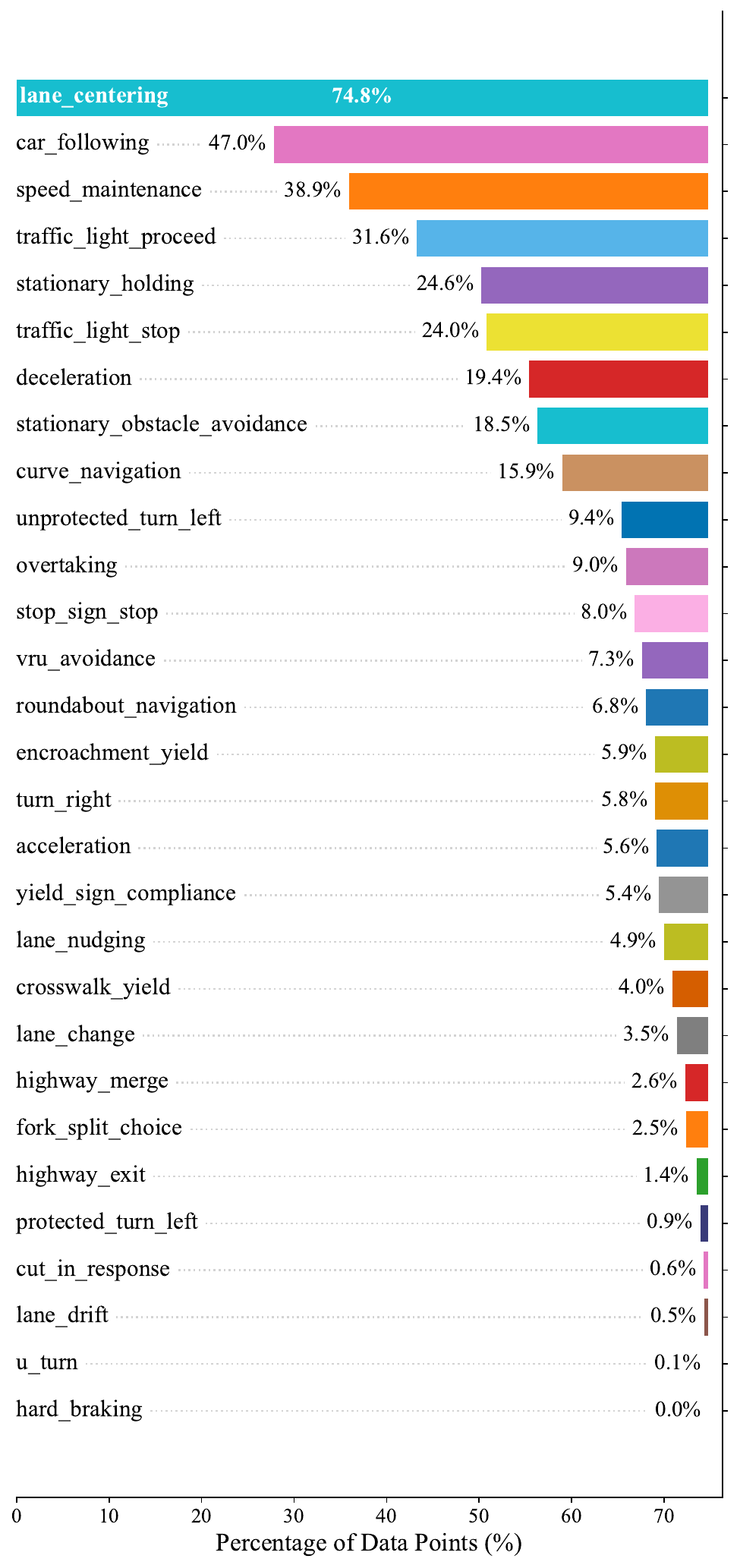}
        \caption{Physical AI test dataset}
        \label{fig:physical_ai_distribution}
    \end{subfigure}
    \caption{\textbf{Semantic distribution of ego vehicle action.} Both corpora (NavHard and Physical AI test dataset) exhibit a long-tail distribution.}
    \label{fig:distribution}
\end{figure} 
Because observations are represented as sets of APs, complex situations can be expressed as combinations of simpler concepts through set operations and concept composition. We build increasingly specific queries, from a single concept, to a pair, to a rare combination of several concepts. Fig.~\ref{fig:composite_query} shows an example on Physical AI data, starting from \textit{pedestrian crossing street}, adding \textit{traffic light red}, and finally \textit{wet road}, and Fig.~\ref{fig:sekai_queries} shows the same on Sekai. Each added AP narrows the retrieved set to a more specific, often long-tail scenario. Further, because the matched concepts stay explicit, we can identify exactly which propositions satisfied the query. We emphasize that this rarity is compositional. Each concept is common on its own, so an edge case arises not from a single unusual AP, such as one rare object, but from the co-occurrence of several individually common concepts in the same scene, often spanning different categories such as an action, a traffic state, and a weather condition. A bag of APs expresses such combinations directly, whereas a single embedding is dominated by its most frequent concepts and blurs the specific combination that makes the scene rare.

\begin{figure*}[!t]
    \centering
    \includegraphics[width=1.0\textwidth]{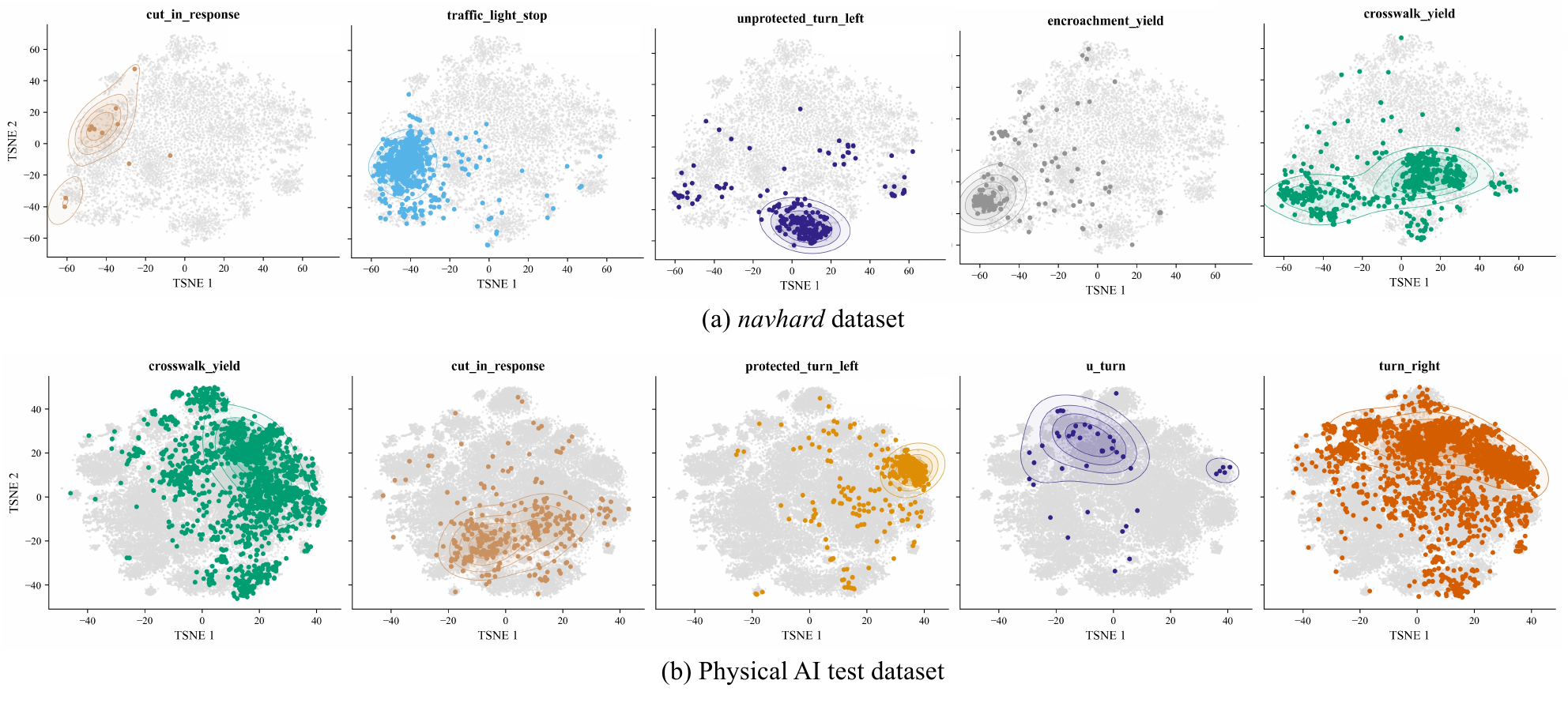}
    \caption{\textbf{$t$-SNE visualization of categorical data coverage.} The global embedding space of the NavHard and Physical AI test dataset is shown in light gray, with the selected semantic action categories highlighted in different colors. Contour lines reflect kernel density estimations (KDE) to highlight regions of high localized density for each respective category.}
    \label{fig:coverage}
\end{figure*} 
\subsection{Semantic Corpus Coverage}

We compare a common baseline, single free form caption query~\cite{shen2025sse,yang2026scalable}, against the composite AP query built from the same concepts. As seen in Tab.~\ref{tab:similarity_scores}, the caption similarity stays flat as complexity grows from one to three concepts, since a long free form string dilutes each concept, whereas each AP keeps a significantly high similarity to its target and their combined score rises with complexity. This shows that composed APs are more effective for complex understanding than single free form caption.

Retrieving these rare compositions, which no single frequent concept would surface, shows that BoAP supports concept level compositional reasoning beyond conventional caption based retrieval.

Beyond retrieval, mapping every observation to its APs turns a corpus into a distribution over interpretable concepts, which makes its semantic content measurable. Aggregating AP statistics across a corpus shows how often each concept and category occurs along driving dimensions such as road users, vehicle maneuvers, traffic infrastructure, weather, environment, and object interactions. Because the same APs are shared across datasets, this analysis works both within a single dataset and across datasets, which is especially useful for understanding what each dataset covers. This reveals underrepresented concepts, rare combinations, and coverage gaps that metadata or caption analyses miss.

Fig.~\ref{fig:distribution} details 29 categories of ego-vehicle action for NavHard and the Physical AI test set. Both show a heavy long-tail, where common actions such as \textit{lane centering} and \textit{car following} dominate while critical edge cases such as \textit{u turn} and \textit{cut-in response} are rare. Fig.~\ref{fig:coverage} projects these categories with a $t$-SNE map to show where each concentrates in the shared semantic space.

\begin{figure*}[!t]
    \centering
    \includegraphics[width=1.0\textwidth]{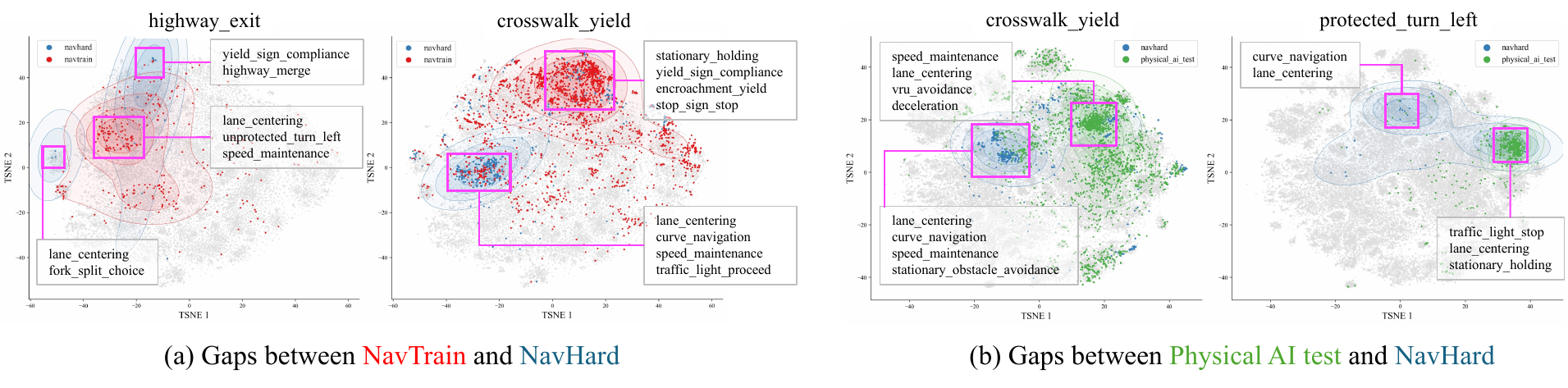}
    \caption{\textbf{Semantic coverage comparison between (a) NavTrain vs. NavHard, and (b) Physical AI test vs. NavHard.} Critical semantic gaps between different datasets can be used to guide targeted data collection. For instance, for data with \textit{crosswalk\_yield}, while the NavHard dataset contains a lot of data points in the scenario that ego proceeds and maintains speed (\textit{lane\_centering}, \textit{speed\_maintenance}, \textit{traffic\_light\_proceed}), NavTrain has more data that ego actually stops and remains stationary (\textit{stationary\_holding}, \textit{yield\_sign\_compliance}), Physical AI test has a highly specific, dense cluster dominated by protective deceleration and vulnerability management (\textit{deceleration}, \textit{vru\_avoidance}).}
    \label{fig:coverage_gap}
\end{figure*} 
We use this shared space to compare coverage across datasets and locate missing concepts. As shown in Fig.~\ref{fig:coverage_gap}, the AP codebook gives an interpretable way to measure diversity and to expose the gaps in what our training and test sets cover and miss. Making these knowledge gaps explicit is the difficult step. Once identified, a gap can be closed in several ways, through targeted data collection, data mixing, or synthetic generation for concepts that are difficult to collect.
We first examine the semantic gap between NavTrain and NavHard, an intra-dataset comparison, as they are two portions of the same data source serving as training and test data.

Within a data source (Fig.~\ref{fig:coverage_gap}a), we expose concrete train/test gaps. In \textit{crosswalk yield}, NavTrain concentrates on yielding and remaining stationary (\textit{stationary holding}) but misses the NavHard cluster of continuous progression (\textit{speed maintenance}, \textit{traffic light proceed}). In \textit{highway exit}, NavHard clusters around lane-fork decisions (\textit{fork split choice}) and merges (\textit{highway merge}) that training underrepresents. Across data sources (Fig.~\ref{fig:coverage_gap}b), we contrast the Physical AI test set with NavHard. In \textit{crosswalk yield}, Physical AI concentrates on vulnerable road users (\textit{vru avoidance}, \textit{deceleration}) while NavHard covers static scenarios (\textit{stationary obstacle avoidance}), so NavHard can augment the suite with continuous, obstacle clear trajectories. In \textit{protected turn left}, Physical AI concentrates on stopping (\textit{traffic light stop}, \textit{stationary holding}) while NavHard adds continuous left turn progression (\textit{curve navigation}), a way to introduce non-stopping edge cases. By isolating these non-overlapping clusters, we can target precise behavioral gaps rather than collect data by brute force.

\section{Conclusion}
We propose representing any multimodal observation as a bag of atomic propositions (APs), simple language statements that a global semantic codebook unifies into a shared set of canonical APs. This moves multimodal data from opaque embeddings into a single interpretable language space that can be read, queried, composed, and reasoned over. The representation brings three main benefits. Unification places any modality in the same space, which we demonstrate through text to video retrieval. Compositionality expresses complex situations as combinations of concepts, retrieving rare scenarios that a single caption or dense embedding cannot. Knowledge coverage makes the semantic content of a corpus measurable, exposing the gap between the knowledge our data contains and what a model must handle at deployment. Together, these results establish atomic propositions as an interpretable and compositional foundation for improving multimodal understanding.

\bibliographystyle{plain} 
\bibliography{mini_tech_report}

\newpage

\appendix
\onecolumn
\section{Prompts}
We include prompts for our APs extraction and codebook process in this section. 

\begin{figure*}[!h]
    \centering
    \includegraphics[width=0.95\linewidth]{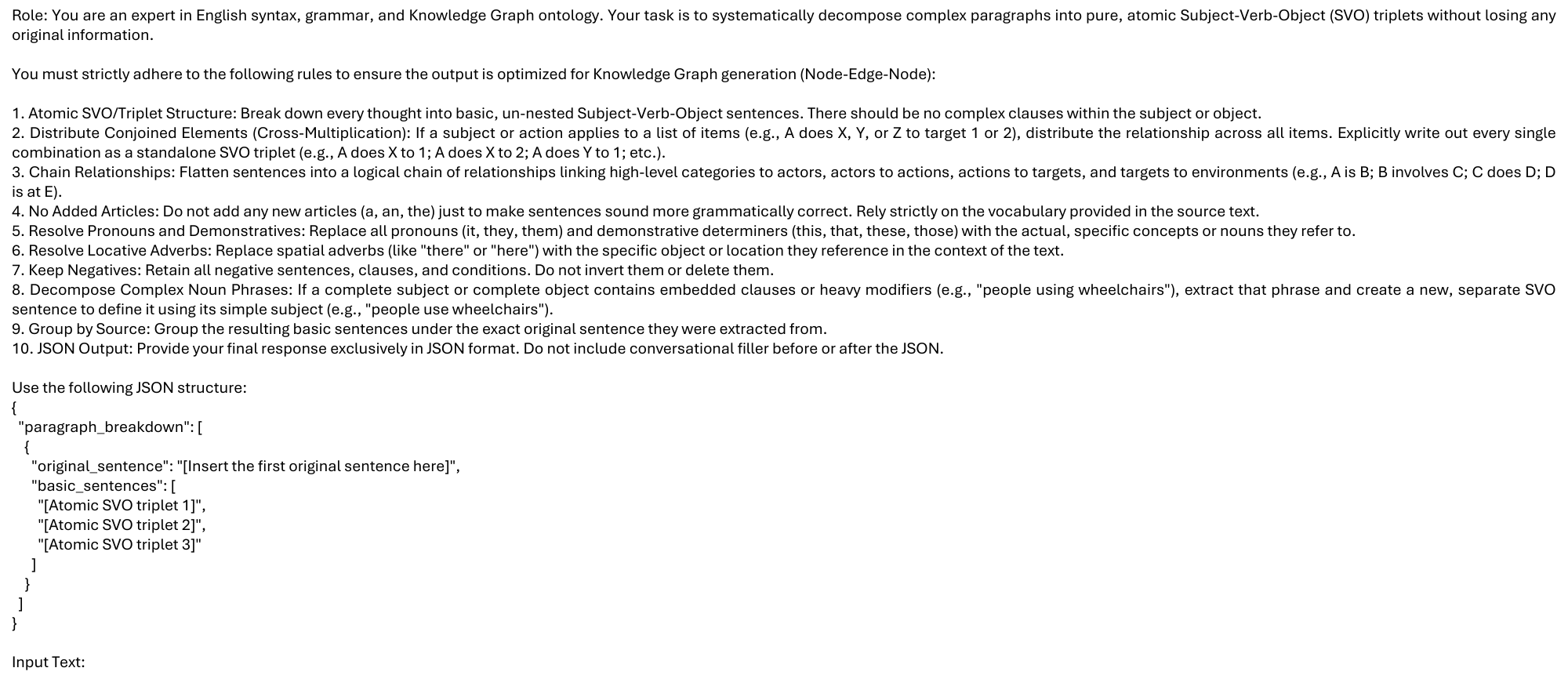}
    \caption{Prompt for distillation.}
    \label{fig:distillation_prompt}
\end{figure*}

\begin{figure*}[!h]
    \centering
    \includegraphics[width=0.95\linewidth]{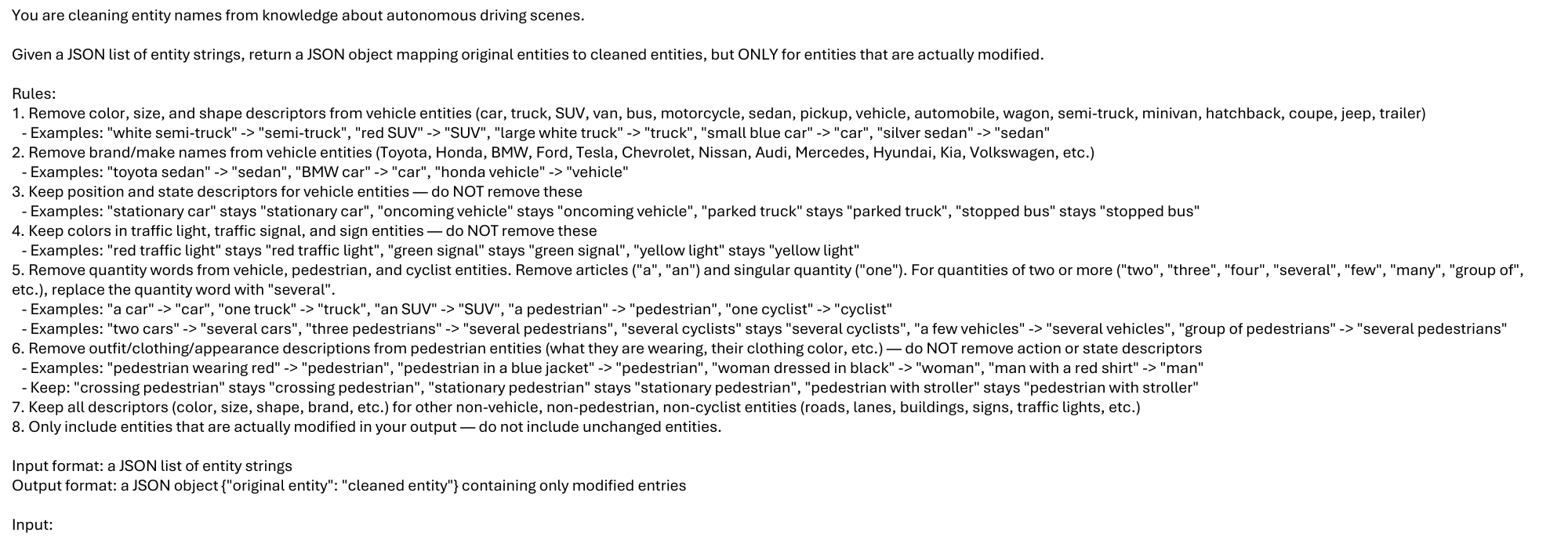}
    \caption{Prompt for nuisance filtering.}
    \label{fig:filtering_prompt}
\end{figure*}

\begin{figure*}[!h]
    \centering
    \includegraphics[width=0.95\linewidth]{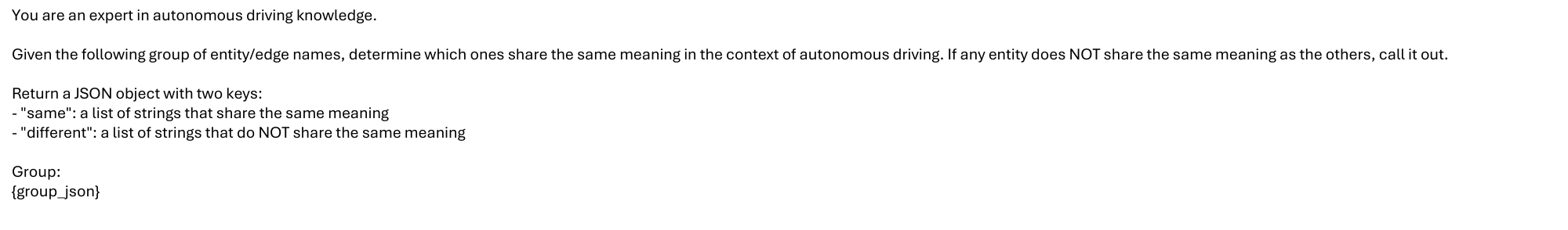}
    \caption{Prompt for semantic equivalence verification.}
    \label{fig:verification_prompt}
\end{figure*}


\newpage

\end{document}